\documentclass[aps,pre,amsmath,amssymb,reprint,superscriptaddress]{revtex4-1}

\usepackage{hyperref}
\hypersetup{colorlinks,}

\usepackage[inkscapelatex=false]{svg}
\usepackage{graphicx,dcolumn,bm}
\usepackage{sidecap}
\usepackage{booktabs,array}
\usepackage{siunitx}
\usepackage{subcaption}

\begin{document}
\title{Data-driven transport modelling without overfit}

\author{Peter Vanya}
\affiliation{Sev.en Global Investments, V Celnici 4, 11000 Prague, Czechia}
\email{peter.vanya@gmail.com}

\author{Katar\'{i}na \v{S}imková}
\affiliation{Vrije Universiteit Brussels, Brussels, Belgium}
\affiliation{Université Libre de Bruxelles, Brussels, Belgium}

\author{Rastislav Farka\v{s}}
\affiliation{Ministry of Finance of the Slovak Republic, Štefanovičova 5, 81782 Bratislava, Slovakia}


\begin{abstract}
Macroscopic transport modelling aims to predict traffic flows after proposed public policy interventions, such as a new road or railway section or a temporary road closure. As such, it is a vital step in infrastructure planning and development. Traditionally, building a transport model has relied on complex understanding of socio-economic characteristics of the population requiring expensive data collection via surveys, which are prone to biases. Previous numerical frameworks to optimize transport models to fit observed traffic flows are not easily-interpretable and can lead to overfit. We present here an alternative: a data-driven modelling protocol with objective function based on traffic counts, which can be nowadays cheaply and reliably obtained; explainable model weights; and a controlled path to increase model complexity and accuracy. We demonstrate our approach on several toy and realistic examples, and suggest ways to generalize to multimodal systems including public transport.
\end{abstract}

\maketitle

\section{Introduction}
A transport model (TM), which predicts demand for road and railway sections, is a critical step in public investment process. Outcomes of transport modelling serve as inputs for cost-benefit analyses supporting decisions, whether an investment should be realized, and which are worth billions of dollars annually in a typical mid-sized country~\cite{tag,Sunstein2018}. A high-quality TM should enable reliable exploration of various investment scenarios and fine-tune the project to achieve the highest economic return on investment for the taxpayer.

Generally, there are two dominant approaches to model traffic. Microscopic models follow movements of separate agents (e.g. a car or a pedestrian), which is important for optimizing transport within cities~\cite{Treiber_book_2013,Helbing_PRE_1995}. On the other end of spectrum there is macroscopic modelling, which aims to deduce traffic flows typically measured in vehicles per 24-hour interval (or veh/24h) between cities, regions or distinct countries.

The latter approach has gained traction with the post-war advent of highways in the Western countries in 1960s and the need to estimate traffic flows \emph{ex ante} for the purpose of planning~\cite{Wilson_book_1974,Dafermos1969}. This work aims to expand on it.

A macroscopic TM is based on graph theory~\cite{Ortuzar_book_2011}. Basic inputs of a macroscopic TM include a set of zones, where people live, which serve as graph vertices, and the network of road or railway links interconnecting these zones, which serves as graph edges. Modelling then consists of four steps~\cite{Ortuzar_book_2011}: (1) trip generation; (2) trip distribution yielding an origin-destination (OD) matrix; (3) mode choice between public and private transport; and (4) assignment of traffic flows to road and railway network.

Despite decades of evolution, there remains an element of art in the science supporting this field; key success metrics are not exactly defined and, consequently, a standardized process to achieve a desired model quality or accuracy is not readily available.

With the advent of big data and machine learning, it should be, in principle, possible to construct high-quality TMs by feeding model more data and fine-tuning hyperparameters~\cite{Bishop_book_2006,Krizhevsky2012}. Such data-driven approach can also shed light on population characteristics, which are nowadays expensively collected via surveys. There is room for higher-quality and, at the same time, cheaper TMs.

Early attempts at data-driven transport modelling date back to early 1980s. Willumsen and van Zuylen Willumsen~\cite{VanZuylen1980,Willumsen1978a,Willumsen1978b,Willumsen1981} developed methods to estimate the OD matrix from observed traffic flows using least-squared minimization or information maximization. However, both of these OD matrix modifications do not preserve the logical flow of the model, which must start at the trip generation step.

Varying OD matrix elements to fit traffic flows implicitly modifies the population characteristics of the investigated region and the mobility patterns of the population, from which the OD matrix was built in the first place and which serve as model inputs. Furthermore, free variation of all $\mathcal O(N_{\text{zones}}^2)$ OD matrix elements to optimize for traffic counts leads to an overfit; traffic counts scale as $\mathcal O(N_{\text{links}})$, while it holds that $N_{\text{links}} \ll N_{\text{zones}}^2$. In practice, traffic counts are even more sparse than road links, as the former are rarely collected on every road section. As a result, the predictive ability of a TM with thus adjusted OD matrix breaks down.

\begin{figure*}[t]
\centering
\includegraphics[width=0.9\textwidth]{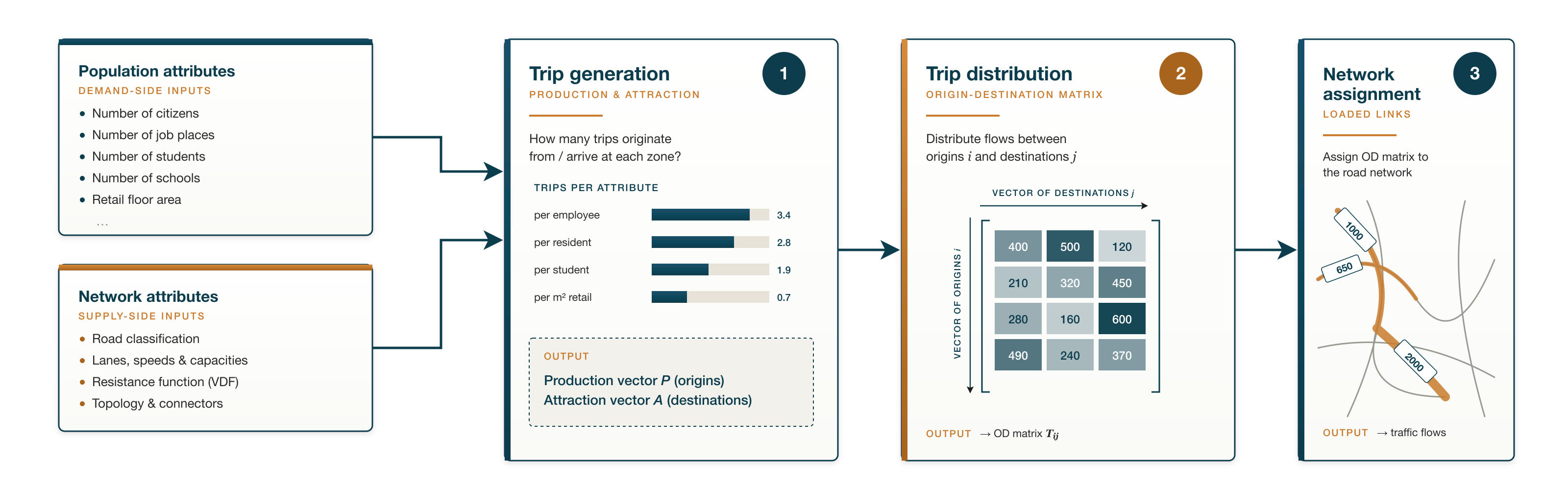}
\caption{A general scheme of a three-step transport model workflow, which includes trip generation, trip distribution and network assignment. This workflow excludes the mode choice between public and private transport, which is a step between distribution and assignment in a four-step model.}
\label{fig:scheme}
\end{figure*}

We present here an alternative data-driven approach, aiming to eliminate overfitting while calibrating the model to make it accurate. This inherently boosts the predictive capabilities of the model when new links are added in line with proposed traffic network changes. Like Refs~\cite{VanZuylen1980,Willumsen1978a,Willumsen1978b,Willumsen1981}, we also start from traffic counts as the most suitable predictor. These can be collected cheaply and on scale, as is already the case for some developed cities~\cite{berlin_counts}. Instead of directly modifying all OD matrix elements, we propagate the objective function deeper, back to the trip generation step (Fig.~\ref{fig:scheme}), and vary only few well-defined weights. There are only two such weights for each trip-generating pair of population attributes, or \emph{demand strata} (DS); hence, their total number grows as $\mathcal O(N(\text{DS}))$, which is significantly lower than the number of observed data samples.

The main advantage of our approach is that the internal logic of the OD matrix is upheld, as its elements depend on only a few independent model weights controlling trip generation and distribution. This is critical to correctly assess the impact of new road bypasses, highway or railway sections, which are among the most capital-intensive projects in any country. Our approach also supports the phenomenon of induced demand, the result increasing trips due to reduced travel times or costs~\cite{Lee1999}. The inferred model weights are explainable, directly comparable to survey data, and transferable to other TMs in comparable regions, as they represent the generic travel patterns of the population.

Section~\ref{sec:protocol} explains the simulation process; Section \ref{sec:results} presents the application to several toy and realistic models.


\section{Data-driven modelling protocol}
\label{sec:protocol}
Having motivated the aims of macroscopic transport modelling, we now proceed with explaining the choice of model features, weights, hyperparameters, and outputs.

\subsection{Inputs}
In transport modelling, there are two input groups that should be functionally separated.

The first group is the geographical reality in the form of road network. This includes roads divided into sections, each with its effective maximum speed\footnote{Maximum effective speed is lower not the same as legal speed, since it is influenced by road width, slope, curvature and the level of surface degradation.}, capacity, and a resistance, or volume-delay (VD) function~\cite{Ortuzar_book_2011,BPR1964} determining the travel time $t$ as a function of traffic flow $Q$:
\begin{equation}
t(Q) = t_0 \left[ 1+\alpha_1 \left( \frac{Q}{Q_{\rm{max}}} \right)^{\alpha_2}\right],
\end{equation}
where $Q_{\max}$ is the road capacity, $t_0$ travel time at zero traffic flow, and $\alpha_1$ and $\alpha_2$ constants~\footnote{As a rule of thumb, $\alpha_1=0.15$ and $\alpha_2=4$ are reasonable estimates.}.

Another group are socio-economic attributes of the population within the examined perimeter, which determine the travel patterns. Inhabitants are clustered into zones serving as graph vertices, which are connected to the road network. Patterns include employees travelling to work, which is contained in zones with job places, or pupils travelling to school, which is in zones with school places. On the simplest and most general level, people travel where other people are, i.e. density is attracting density, in an analogy to Coulomb or gravitational laws from physics.

\subsection{Traffic model and its weights}
Each such trip-generating travel pattern defines a \emph{demand stratum} (DS), or a pair of trip production and attraction attributes with associated \emph{mobility}, i.e. number of vehicle trips per day (henceforth denoted as $\mu$). Model complexity can be  naturally controlled by the number of DS; an expansion of the set of population attributes should lead to more refined and accurate transport models.

After trip generation, trips are distributed between zones via so-called gravity models, which derive from the maximum entropy principle~\cite{Jaynes1957,Wilson_book_1970}. The distribution of trips has the form of the origin-destination (OD) matrix $T_{ij}$ for each DS. Starting from a vector of origins $O_i$, a vector of destinations $D_j$ and a cost or deterrence matrix $C_{ij}$, which is determined most often by travel time $t_{ij}$ between zones as graph vertices, the number of trips $T_{ij}$ between zone $i$ and zone $j$ is:
\begin{equation}
T_{ij} \sim \frac{O_i D_j}{f(C_{ij})}.
\end{equation}
Here, $f(x)$ is a generalized deterrence function, commonly an exponential or a polynomial:
\begin{equation}
f(c) = \begin{cases}
e^{\beta c}, \\
c^{\beta},\\
\end{cases}
\label{eq:deterrence}
\end{equation}
where $\beta$ is a weight controlling how far people are willing to travel on average for a given purpose (job, school, leisure, holiday etc.). $\beta$ is closely related to the mean trip length, which can be statistically measured via surveys. Generally, it follows that wealthier societies or demand strata are willing to travel further.

As a final step, matrix elements $T_{ij}$ are assigned to the road network for each DS, yielding observable traffic flows. This process can be one-off or iterative to account for gradual filling of the roads, cycling between assignment, cost matrix $C_{ij}$ and OD matrix.

In summary, three-step transport modelling proceeds as follows: (i) trip generation by defining DS and associated mobilities $\mu$; (ii) trip distribution to obtain OD matrix for each DS; (iii) network assignment to generate the traffic flows on the road network~\footnote{Compared to the general macroscopic transport-modelling framework as described in the Introduction, the step of mode choice is missing, as this framework is focused on private-car traffic for now.}.

With this definition the modelling workflow, we now select two principal groups model weights for each DS. The first is \emph{mobility} $\mu$ defined as the number of trips per person per 24h period~\footnote{To convert between persons to passenger vehicles on the road, person-trips must be divided by average vehicle occupancy, typically between 1 and 2 persons/veh.} Traditionally, mobility patterns have been collected via surveys involving individual travel logging, which are expensive and prone to systematic biases. The second is \emph{the distribution parameter} $\beta$, which appears in the gravity trip distribution model and controls for the mean trip length, which can also be collected by surveys.

\subsection{Output and objective function}
The most straightforward output is the traffic flows from the assignment step. These can be compared directly with data from automatic traffic counters, which are nowadays collected cheaply, reliably and at scale from camera streams~\cite{Tang2022,Song2019}. These characteristics position the traffic flows as the most suitable model quality predictors from pragmatic and data-driven point of view.

A typical error measure in transport modelling is the GEH statistic with the following relation between predicted hourly $P_i$ and measured $M_i$ traffic flows on any given road section:
\begin{equation}
\text{GEH}\text{ (hourly)} = 
\sqrt{\frac{2(P - M)^2}{P + M}}.
\end{equation}

As a rule of thumb, daily flows are roughly 8-12x larger than typical hourly flows. Selecting a middle value of 10x, the error for daily flows becomes:
\begin{equation}
    \text{GEH}\text{ (daily)} = \sqrt{10} \text{ GEH}\text{ (hourly)}
\end{equation}
In the subsequent discussion, we will use the hourly GEH as a default metric for consistency with existing literature.

The GEH error is more suitable than a simple RMS error, since the former accounts for the scale of the underlying quantity. Traffic flows in a typical country are spread across at least two orders of magnitude (between 1k and 100k veh/24h), and the simple absolute difference cannot reasonably capture this wide spread.

The objective function $J$ is an average GEH error for $N_{\text m}$ road sections with observed traffic flows: $J = \sum_i \text{GEH}_i/ N_{\text m}$.

According to the British Guidelines WebTAG, a well-calibrated complex TM is expected to have at least 85\% of hourly traffic counts with GEH $< 5.0$~\footnote{tag,Friedrich2019}. In qualitative terms, an average error around or slightly above this value in case of a simple TM should lead to satisfactory performance and inputs for a final investment decision.

\subsection{Model optimization}
Having defined model weights, output, and the objective function, it is now possible to perform machine learning to calibrate the model to minimize the function with respect to the weights. Such automated procedure enables to avoid the laborious manual calibration and unlocks human resources for tasks with more consequential impact on model quality. Such learned weights are now transferrable between region and even between countries with similar level of development.

A minor efficiency barrier is that it is impossible to analytically compute the gradient of the objective function, since the modelling protocol contains the iterative entropy maximization step. In this case, suitable optimization methods are simulated annealing~\cite{Kirkpatrick1983, Xiang1997} or the Nelder-Mead algorithm~\cite{Nelder1965}. A numerical gradient computation was briefly explored as well, but did not lead to satisfactory convergence.


\begin{figure}[t]
    \centering
    \includegraphics[width=0.75\linewidth]{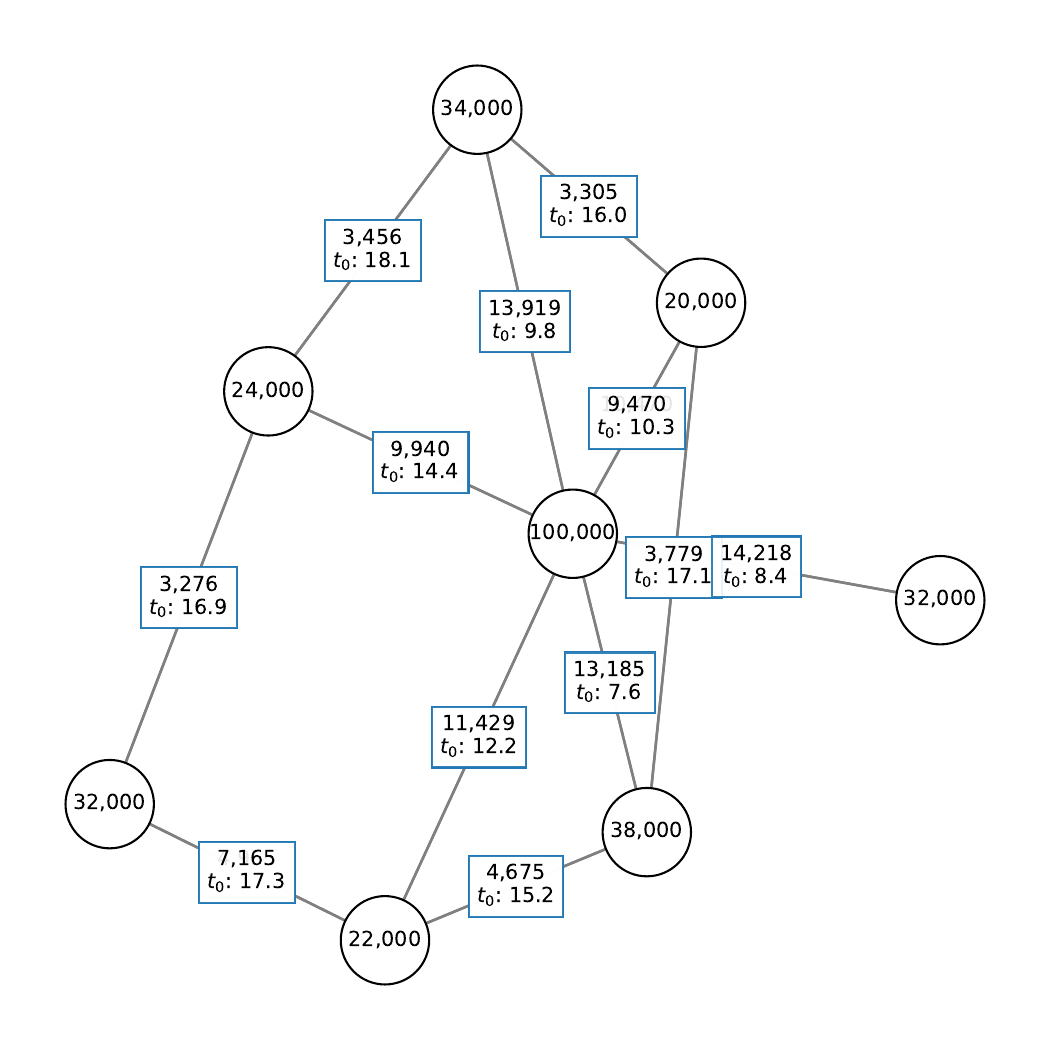}
    \caption{A scheme of a toy model with 8 zones. Zonal populations are in circles; sectional traffic flows (veh/24h) and free-flow travel times (minutes) are in blue boxes.}
    \label{fig:toy-scheme}
\end{figure}

\begin{figure}[t]
    \centering
    \includegraphics[width=0.99\linewidth]{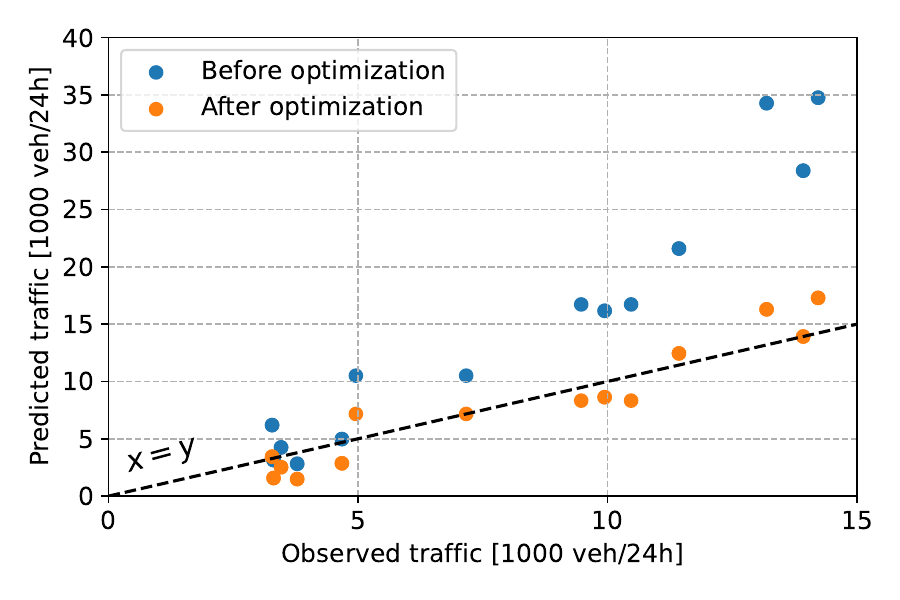}
    \caption{A scatter plot of observed vs predicted traffic flows for a toy model.}
    \label{fig:toy-scatter}
\end{figure}

\section{Applications of data-driven modelling}
\label{sec:results}
After the explanation of the modelling protocol in Section~\ref{sec:protocol}, we now apply it to several transport systems. We start with a simple toy model of only eight zones with size as the only population attribute and demonstrate that the optimization improves the precision. Consequently, we proceed with a realistic regional TM. In all cases, we exploit an open-source Python library TrafficFlow~\footnote{Available at \url{https://github.com/petervanya/traffic-flow}}.

\subsection{Small toy model}
\label{sec:toy}
As a proof of concept, we build a toy model with one large central zone with population of 100k and seven smaller zones with populations between 20k and 40k (Fig.~\ref{fig:toy-scheme}). In the trip distribution step, we use an exponential deterrence function (Eq.~\eqref{eq:deterrence}). A trial run with estimated mobility $\mu=1.5$ (typical number of trips per vehicle in a day) and $\beta=0.1$ yields an average GEH error of 18.1, which is several times higher than common limits for model precision. Running an optimization process using both dual annealing and the Nelder-Mead method decreases the objective function from \emph{c.a.} 14 to only 6.15, with learned weights $\mu=0.70, \beta=0.074$. The scatter plot in Fig.~\ref{fig:toy-scatter} shows a marked improvement upon the first random guess and so demonstrates the viability of the optimization protocol.

\subsection{A realistic regional model}
\label{sec:i51}
Having demonstrated the viability of data-driven optimization on a toy example, we proceed with a realistic model of one region (NUTS 3 in EU nomenclature~\footnote{\url{https://ec.europa.eu/eurostat/web/nuts}}). We consider the Nitra region in western Slovakia, specifically covering a 40$\times$40~km footprint containing road I/51. Here, several infrastructure projects in the form of bypasses are considered in order to speed up travel times between major regional hubs and relieve the municipalities on the main road from excessive transiting traffic and pollution. The dedicated transport model contains 160 zones and 2100 directed road links describing 1050 roads.

The observed traffic flows were taken from the National Traffic Survey from 2015 modified by appropriate growth coefficients according to local traffic norms~\footnote{\url{https://www.ssc.sk/sk/cinnosti/rozvoj-cestnej-siete/dopravne-inzinierstvo/celostatne-scitanie-dopravy-v-roku-2015.ssc}} In total, there were \emph{c.a.} 250 observed traffic flows. It is  theoretically possible to add tens of DS without risking the overfit and thus improve the predictive accuracy of the model.

\begin{figure}[t]
    \centering
    \includegraphics[width=0.99\linewidth]{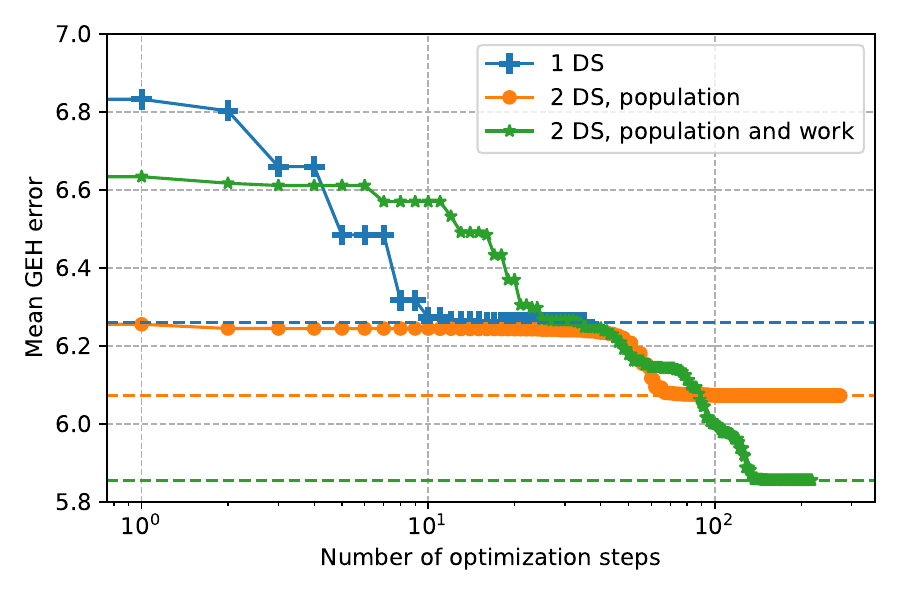}
    \caption{Minimization of the regional model for various configurations of demand strata (DS) using the Nelder-Mead method.}
    \label{fig:i51-curves}
\end{figure}

\begin{figure}[t!]
    \centering
    \includegraphics[width=0.99\linewidth]{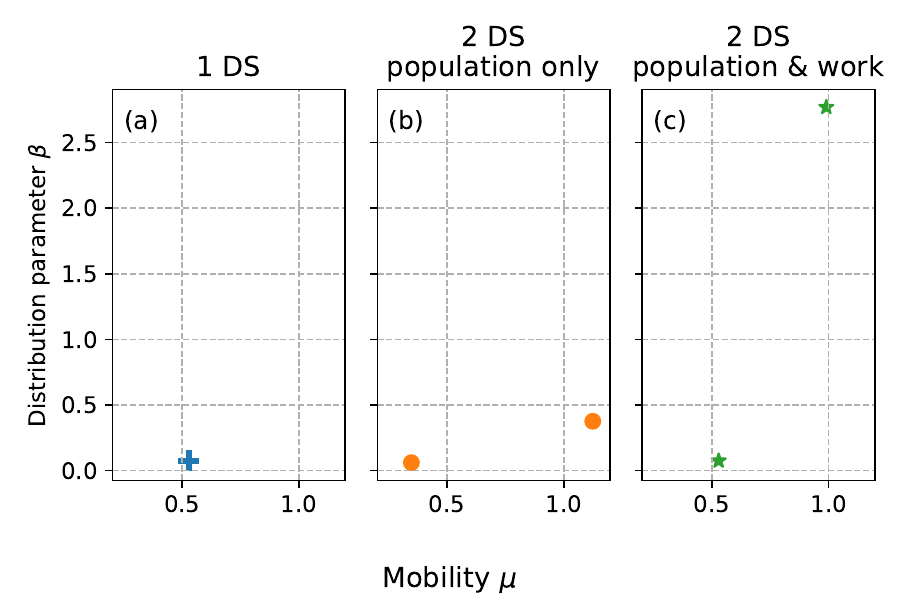}
    \caption{Minimized model weights $\mu$ and $\beta$ for several model complexities characterized by the number of demand strata (DS). (b) Defining two population-to-population DS leads to a new equilibrium with different weights. (c) Adding a population-to-work DS to population-to-population DS preserves one set of weights and creates a new one with greater $\beta$ (bottom right), meaning shorter travel times.}
    \label{fig:i51-params}
\end{figure}

Firstly, a minimal case of a single DS based on population-to-population relation is optimized with the Nelder-Mead method, leading to an error of 6.26, an improvement from an initial guess of weights ($\mu=1, \beta=0.1$) yielding $\text{GEH}=7.7$.

Generally, the consensus in the transport modelling community is that GEH error should be below 5.0 for 60\% to 85\% of transport links~\cite{tag,Ortuzar_book_2011}. However, generally the model complexity controlled by the number of DS and transport modes is not specified in the standard transport guidelines. In this context, the error of 6.26 and 54\% of links with $\text{GEH}\leq5.0$ is an acceptable result for a minimal model with a single DS. Furthermore, this approach of automated determination of optimal weights helps to avoid tedious manual calibration and to direct focus to more tasks with greater impact, such as data collection.

Adding another DS with population-to-population relation leads to further improvement of the error to 6.07 and confirms that the progressive increase in model complexity decreases the error (Fig.~\ref{fig:i51-curves}). The pairs of weights have settled as follows: one DS has lower mobility and smaller $\beta$ implying longer trip lengths, and another DS has higher mobility 1.1 and larger $\beta$, meaning more everyday but shorter trips.

Finally, a two DS configuration with a ``work" attribute is explored, which represents the number of job places in a zone attracting all of the population. The number of jobs in a zone is defined here analytically, with the assumption that only zones with population above a certain cutoff, 5000, contain work places, and smaller zones are residential only:
\begin{equation}
n_{\text{jobs}} = \begin{cases}
\sqrt{x^2 - 5000^2},& x>=5000 \\
1, & x < 5000\\
\end{cases}
\end{equation}

The resulting learned weights (Fig.~\ref{fig:i51-params}) show that the first DS based on population attraction is preserved after optimization, while the new, work-based DS leads to higher mobility and larger $\beta$, implying everyday work-related trips at short distance. This is in line with practical expectations.

\begin{figure}
    \centering
    \includegraphics[width=0.95\linewidth]{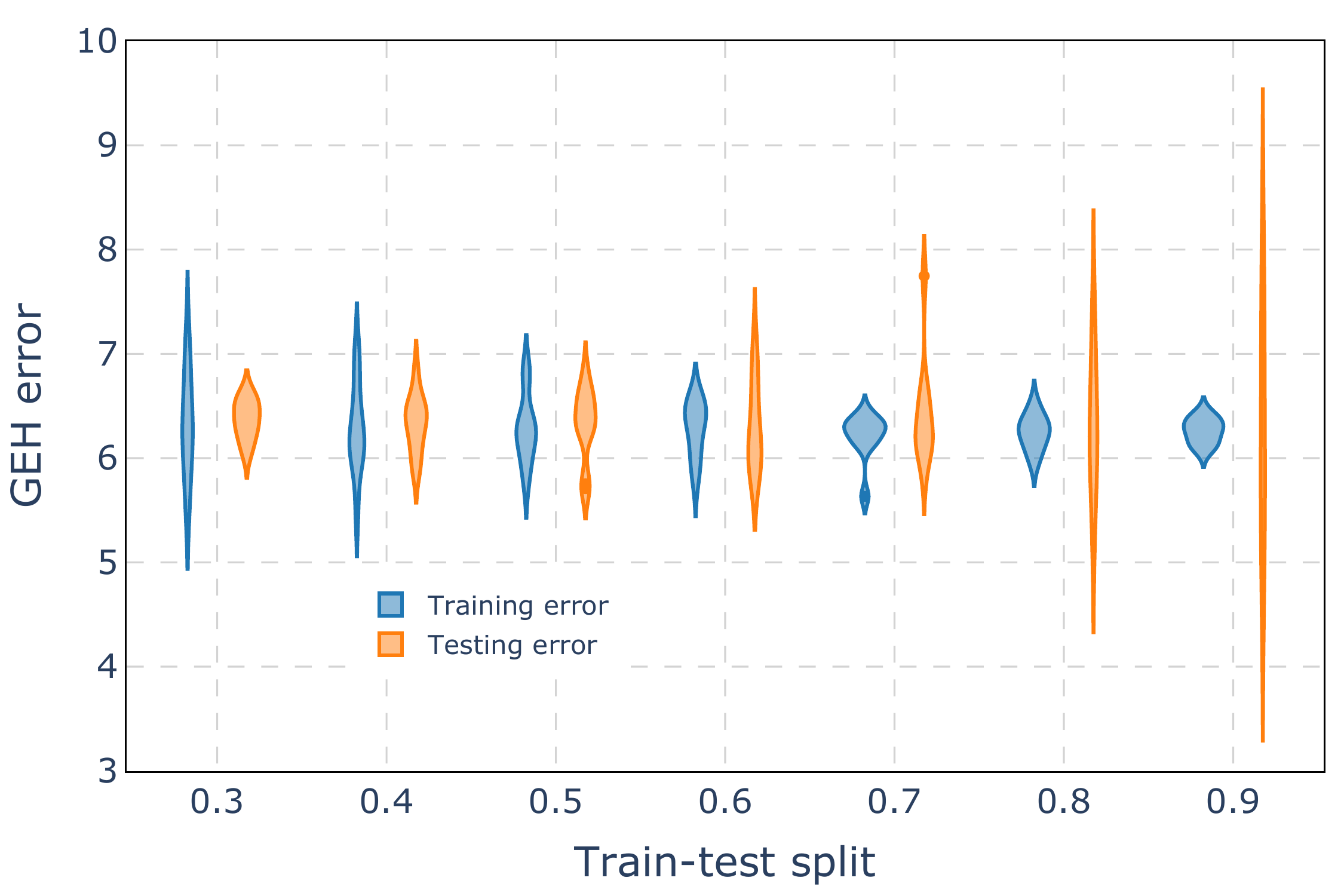}
    \caption{Training and testing error (GEH statistic) as a function of the train-test split for a regional model with 250 total observed traffic flows.}
    \label{fig:traintestsplit}
\end{figure}

To demonstrate the robustness of this modelling protocol to overfitting, we explore a wide range of train-test splits ranging between 30\% and 90\%. Starting with the simplest model with 1 DS, 10 optimizations differing by random seed were performed. The results (Fig.~\ref{fig:traintestsplit}) show that the mean GEH error is already captured by using only 30\% measurements for training, and the variance decreases with rising training split. This agrees with good practice.

We note that it is possible to further reduce model error by improving on transport modelling hyperparameters in the form of the geographic coverage of the simulated region. Firstly, in this regional model not every small village is represented for the sake of simplicity, even if these zones also contribute towards observed traffic flows; secondly, the coverage of larger zones (cities) outside the modelled region, which influence the traffic flows inside, can be refined; thirdly, larger zones, especially cities, within the region can be fine-grained into smaller zones, i.e. city neighborhoods.

In terms of input data scope, for this purpose only total population was recovered from official databases. In principle, job places, school places and other key attraction attributes can can recovered from government databases or open data sources and a much more precise picture about population flows can be derived, subject to legal or inherent limitations to data collection~\cite{Sulikova2024}.
\newline

\section{Conclusions and Outlook}

In this work, we have presented a data-driven transport modelling pipeline with the aim to facilitate the decision-making on large capital-heavy infrastructure projects. The pipeline is built on machine learning from observed traffic flows, whose measurement is nowadays accurate and affordable. The model weights to learn are explainable in a straightforward manner and can be compared with survey data and transferred between models. A clear path for increasing model complexity and accuracy was outlined via adding population attributes. Thus defined modelling process does not interfere with the OD matrix computation containing directional traffic flows, does not overfit even for lower numbers of observed traffic counts and so preserves the predictive ability.

The approach presented here is limited to only one mode of transport, as represented by a three-step model. An obvious next step is to generalize this pipeline to include multiple transport systems, such as freight transport in various forms, or account for public transport, a paradigm known as a four-step model~\cite{Ortuzar_book_2011}. This motivates the addition several extra model weights representing the discrete choice between public and private transport~\cite{BenAkivaLerman_book_1985,BenAkivaBierlaire1999}, a field which has already been a guided by data-driven analysis for several decades. While it has been generally more difficult to collect passenger numbers in public transport, gradual spread of modern sensors and computer vision techniques promises high-quality data for future model development. In light of current environmental and social pressures, it is critical that robust, rigorous and data-driven pipelines are developed to support sound investment decisions.

\bibliography{ref}

\end{document}